\documentclass{article}
\usepackage{arxiv}

\usepackage{tabularx}
\usepackage{multirow}
\usepackage{array}
\usepackage[ruled,vlined]{algorithm2e}

\usepackage[dvipsnames]{xcolor}
\usepackage{censor}
\usepackage[utf8]{inputenc} 
\usepackage[T1]{fontenc}    
\usepackage{hyperref} 
\usepackage{scalerel,stackengine}
\usepackage{ragged2e}
\stackMath
\newcommand\reallywidehat[1]{%
\savestack{\tmpbox}{\stretchto{%
  \scaleto{%
    \scalerel*[\widthof{\ensuremath{#1}}]{\kern-.6pt\bigwedge\kern-.6pt}%
    {\rule[-\textheight/2]{1ex}{\textheight}}
  }{\textheight}%
}{0.5ex}}%
\stackon[1pt]{#1}{\tmpbox}%
}

\usepackage{graphicx}
\usepackage[export]{adjustbox}
\usepackage{colortbl}
\usepackage{paralist}
\usepackage{bbm}
\usepackage{url}            
\usepackage{booktabs}       
\usepackage{amsfonts}       
\usepackage{nicefrac}       
\usepackage{microtype}      
\usepackage{lipsum}	
\usepackage{pifont}

\usepackage{subcaption}
\usepackage{tikz}
\usetikzlibrary{shapes.geometric}
\newcommand{\warningsign}{\tikz[baseline=-.75ex] \node[shape=regular polygon, regular polygon sides=3, inner sep=0pt, draw, thick] {\textbf{!}};}
\usepackage{relsize}
\usepackage{mathtools}

\usepackage{multirow}
\usepackage{multicol}
\usepackage{paralist}
\usepackage{float}
\usepackage[numbers]{natbib}

\title{ARTICLE: Annotator Reliability Through In-Context Learning}

\author{
Sujan Dutta\\
  \small{Rochester Institute of Technology}\\
  \texttt{sd2516@rit.edu}\\
\And
  Deepak Pandita\\
  \small{Rochester Institute of Technology}\\
  \texttt{deepak@mail.rit.edu} \\
\And
Tharindu Cyril Weerasooriya \\
  \small{Rochester Institute of Technology}\\
  \texttt{cyril@mail.rit.edu} \\
\And
 Marcos Zampieri\\
 \small{George Mason University}\\
 \texttt{mzampier@gmu.edu}\\
\And
 Christopher M Homan\\
 \small{Rochester Institute of Technology}\\
 \texttt{cmhvcs@rit.edu}\\
\And
Ashiqur R. KhudaBukhsh\thanks{Ashiqur R. KhudaBukhsh is the corresponding author.} \\
  \small{Rochester Institute of Technology}\\
  \texttt{axkvse@rit.edu} \\
}

\newcommand{\Dv}{$\mathcal{D}_\texttt{VOICED}$}
\newcommand{\Dtr}{$\mathcal{D}_\texttt{TR}$}
\begin{document}
\maketitle

\begin{abstract}
\textcolor{red}{\warningsign This paper discusses and contains content that is offensive or disturbing.} 

Ensuring annotator quality in training and evaluation data is a key piece of machine learning in NLP. Tasks such as sentiment analysis and offensive speech detection are intrinsically subjective, creating a challenging scenario for traditional quality assessment approaches because it is hard to distinguish disagreement due to poor work from that due to differences of opinions between sincere annotators. With the goal of increasing diverse perspectives in annotation while ensuring consistency, we propose \texttt{ARTICLE}, an in-context learning (ICL) framework to estimate annotation quality through self-consistency. We evaluate this framework on two offensive speech datasets using multiple LLMs and compare its performance with traditional methods. Our findings indicate that \texttt{ARTICLE} can be used as a robust method for identifying reliable annotators, hence improving data quality.

\end{abstract}


%

\section{Introduction}

From classical supervised systems~\cite{carbonell1983overview} to the RLHF framework~\cite{christiano2017deep}, human input plays a central role in human value-aligned AI and NLP systems. Crowdsourcing is a well-studied, affordable, and distributed framework that allows data collection from broad and diverse annotator pools within a short period~\cite{gray_ghost_2019,kahneman_noise_2021,wang2013perspectives}. The benefits of crowdsourcing notwithstanding, enforcing quality control and estimating annotation quality remain a long-standing challenge~\cite{lease2011quality,huang_incorporating_2023}. 

Conventional approaches to distinguish \textit{high} 
from \textit{poor} quality annotators are typically based on outlier detection, where the divergence from aggregate opinions is considered a signal of poor quality annotation~\cite{dumitrache2018crowdtruth,leonardelli_agreeing_2021,davani_dealing_2022,ustalov_learning_2024}. However, for subjective tasks~\cite{passonneau2012multiplicity,pavlick2019inherent,uma_learning_2021,nie2020can,jiang2022investigating,deng-etal-2023-annotate}, such outlier-based approaches can potentially muffle minority or unique perspectives, leading to annotation echo chambers. Consider a war corpus where annotators hail from countries $\mathcal{A}$  and $\mathcal{B}$. Even simple questions like \textit{who is winning the war} could have drastically different responses depending on which country the annotator belongs to. If a pool has an overwhelming presence of $\mathcal{A}$, any perspective that annotators from $\mathcal{B}$ could contribute to will be eliminated since their responses will be visibly different from the majority view. 

\begin{figure*}[htb]
    \centering
    \includegraphics[width=\textwidth]{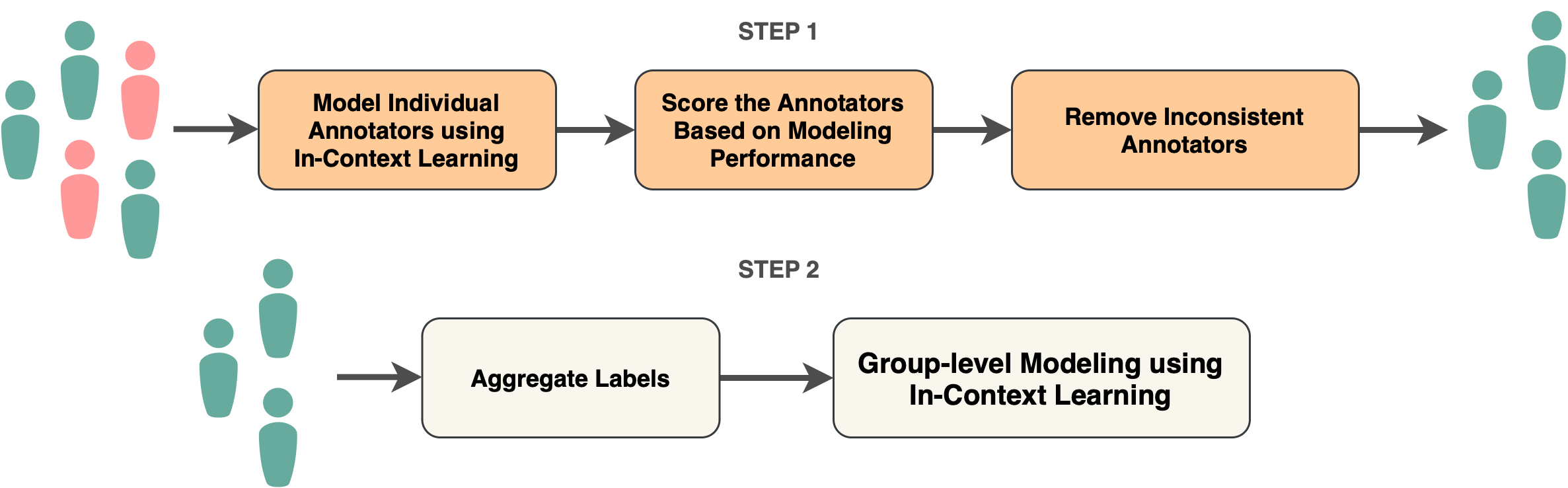}
    \caption{Schematic Diagram of \texttt{ARTICLE}.}
    \label{fig:article}
\end{figure*}

This paper introduces an alternative path to estimate annotator quality through the lens of self-consistency. Prior work in this domain explored to address it through the lens of annotation patterns of individual annotators \citep{dawid_maximum_1979,hovy_learning_2013,ustalov_learning_2024}, without taking into account what is being annotated (context) and information of the annotator. Suppose we are interested in collecting a dataset of offensive speech. If we observe that a given annotator has marked one instance that attacks an ethnic group as highly offensive while marking another instance with an even sharper attack on the same group as not offensive, we immediately know that the annotator's responses are not self-consistent. Incorporating self-consistency into the annotation quality estimation process has the following benefits. First, it bypasses the requirement of having annotations from multiple other annotators to compute divergence from aggregate opinion, thus promising to be more resource-efficient. Second, this approach preserves unique but self-consistent perspectives, which outlier-based methods might eliminate.

While the notion of self-consistency has been applied to diverse settings (see, e.g.,~\citet{DBLP:conf/iclr/0002WSLCNCZ23,cooper2024arbitrariness}), to our knowledge, this paper first applies self-consistency for rater quality estimation on subjective annotation tasks. The introduction of large language models (LLM) with larger context lengths for language understanding has also led to research on utilizing LLMs \citep{gilardi_chatgpt_2023,he_if_2024} as human annotators. However, prior research has focused on using the LLM \citep{he_if_2024} to replace the majority opinion of data annotation but not the intricate annotator-level labels.


\paragraph{Contributions.} Our contributions are the following:
\begin{enumerate}
\item We introduce \texttt{ARTICLE}, a novel framework to estimate annotator quality through self-consistency;  
\item We evaluate this framework on two well-known English offensive speech datasets, (1) \texttt{Toxicity Ratings} \cite{kumar2021designing} henceforth \Dtr~and (2) \texttt{VOICED} \cite{weerasooriya2023vicarious} henceforth \Dv~and we contrast our approach with CrowdTruth (CT) \cite{dumitrache2018crowdtruth}. 
\end{enumerate}



\section{Related Work}


Crowdsourcing platforms such as Amazon Mechanical Turk, Toloka, and Prolific have played a critical role over the years for collecting annotations for training models \cite{kahneman_noise_2021}. However, just as with any task with human annotators in the loop, prior research has identified instances when annotators have been inconsistent with providing information \cite{huang_incorporating_2023, abercrombie2023temporal}. Röttger \textit{et al.} \cite{rottger2021two} demonstrated the impact of the paradigm (subjective or prescriptive) used during the survey on the (dis)agreement level of the annotations. These characteristics have led to research for modeling annotators and rating them for reliability. 

\citet{dawid_maximum_1979} presented the initial two-stage generative model for inferring ground truth from unreliable annotators. The model assumes each annotator has a concealed error rate and utilizes expectation maximization to iteratively estimate these error rates along with the most probable ground truth labels based on the current error rate estimates. \citet{hovy_learning_2013} extended this model with a bipartite annotator model that distinguishes between spammers and non-spammers. CrowdTruth \cite{dumitrache_crowdtruth_2018} is another method for measuring reliability of the annotators and the entire dataset as a whole based on their overall agree-ability with other annotators.  

However, a limitation of prior work is not taking into consideration the content of the data item that is being annotated for scoring the performance of the annotators and how consistent the annotator is in-terms of annotating \citep{cooper_arbitrariness_2024}.  In our research we explore how to utilize the capabilities of the LLM for understanding and identifying inconsistencies of the annotators utilizing context of the annotation task. 


\section{Methodology}
\begin{figure*}[htb]
    \centering
    \includegraphics[width=0.3\textwidth]{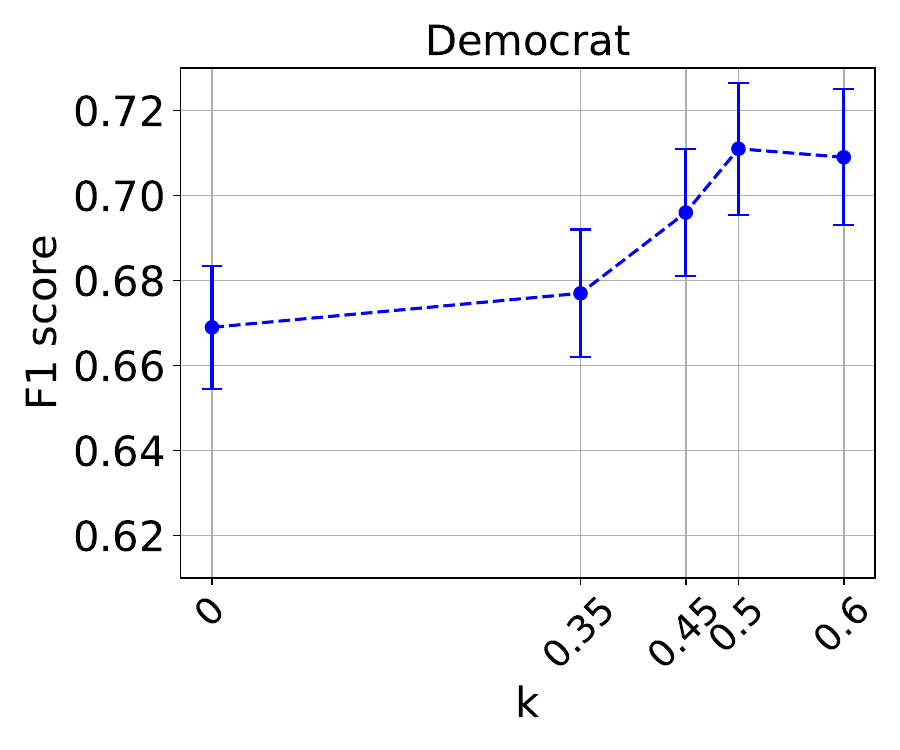}
    \hspace{.5cm}
    \includegraphics[width=0.3\textwidth]{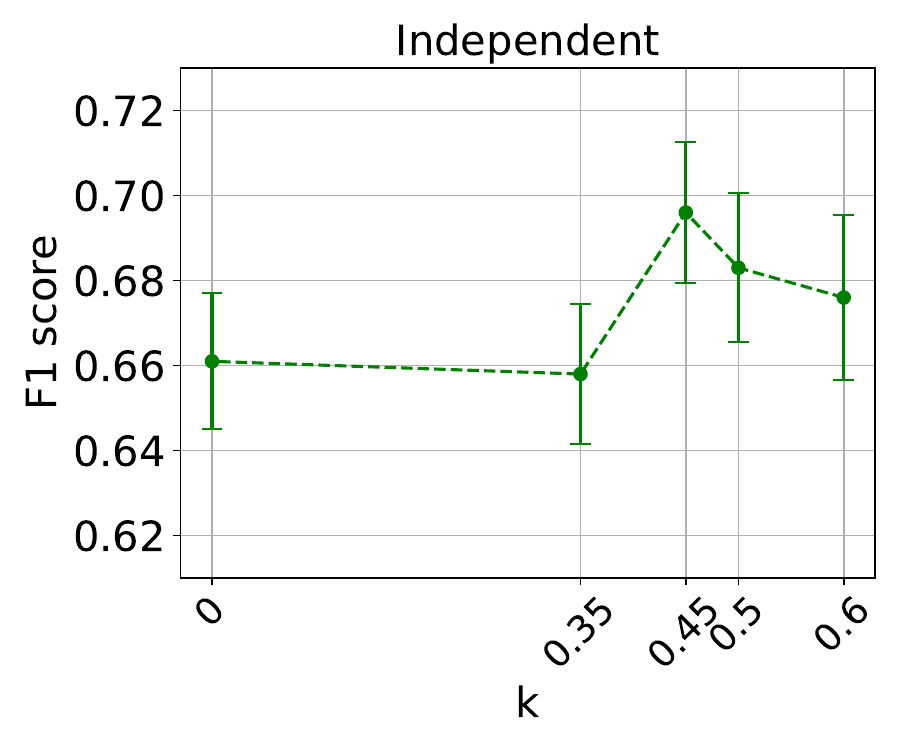} 
    \hspace{.5cm}
    \includegraphics[width=0.3\textwidth]{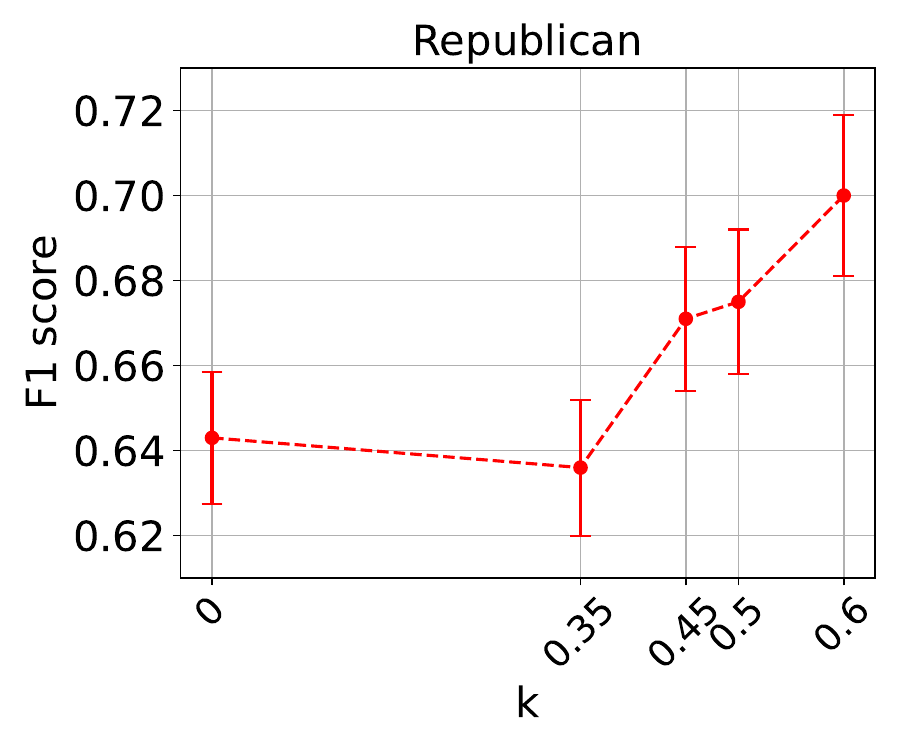}
    \caption{Group-level model performance at different $k$ values in \Dtr. The error bars indicate 95\% confidence interval.}
    \label{fig:f1_tr}
\end{figure*}
\begin{figure*}[htb]
    \centering
    \includegraphics[width=0.3\textwidth]{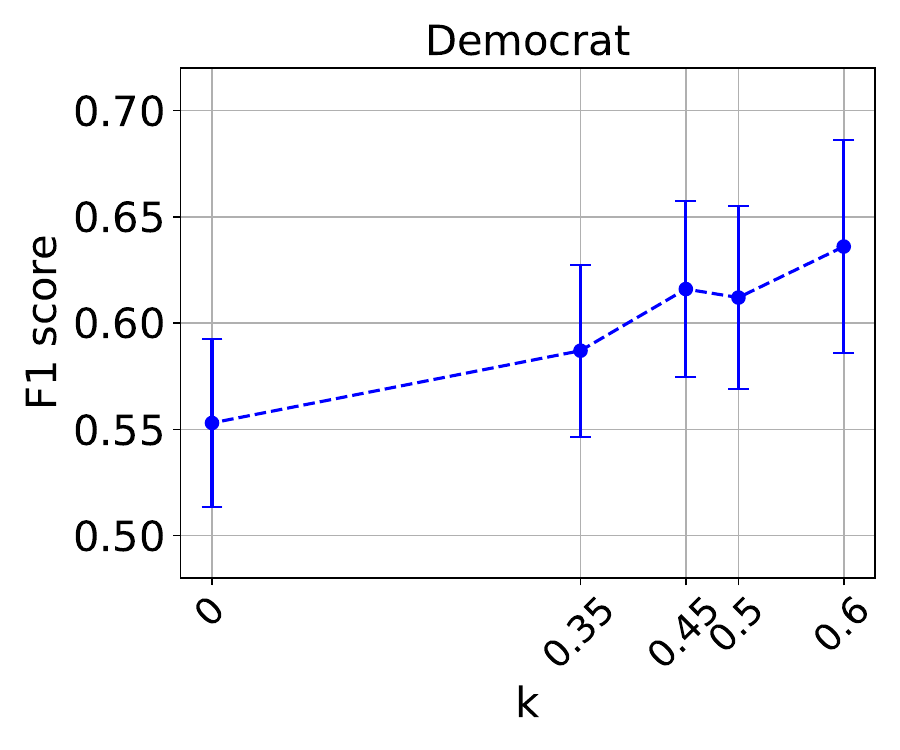}
    \hspace{.5cm}
    \includegraphics[width=0.3\textwidth]{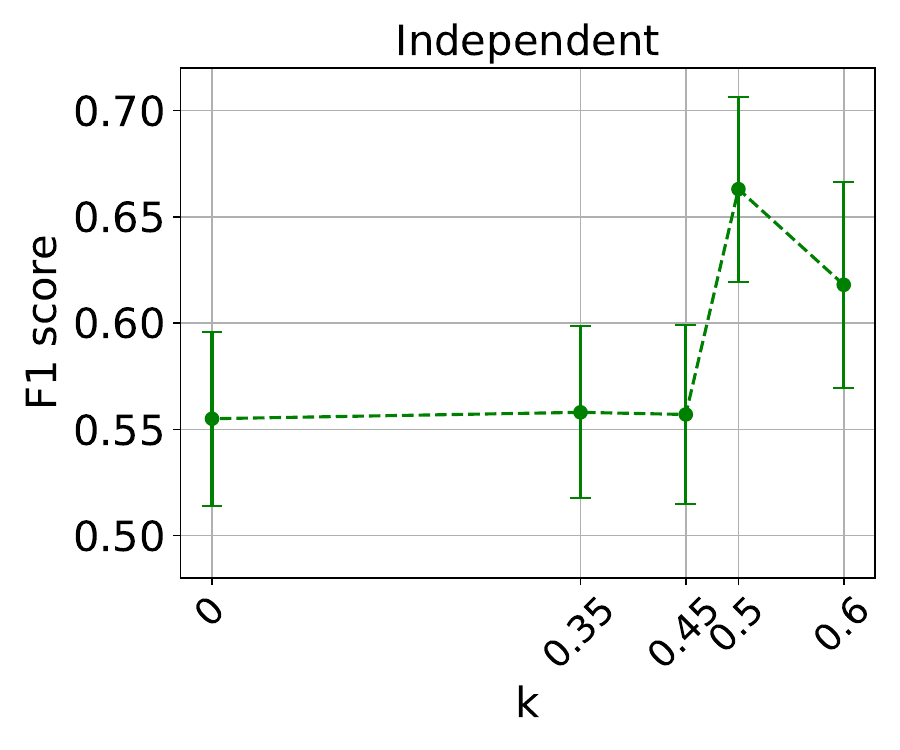} 
    \hspace{.5cm}
    \includegraphics[width=0.3\textwidth]{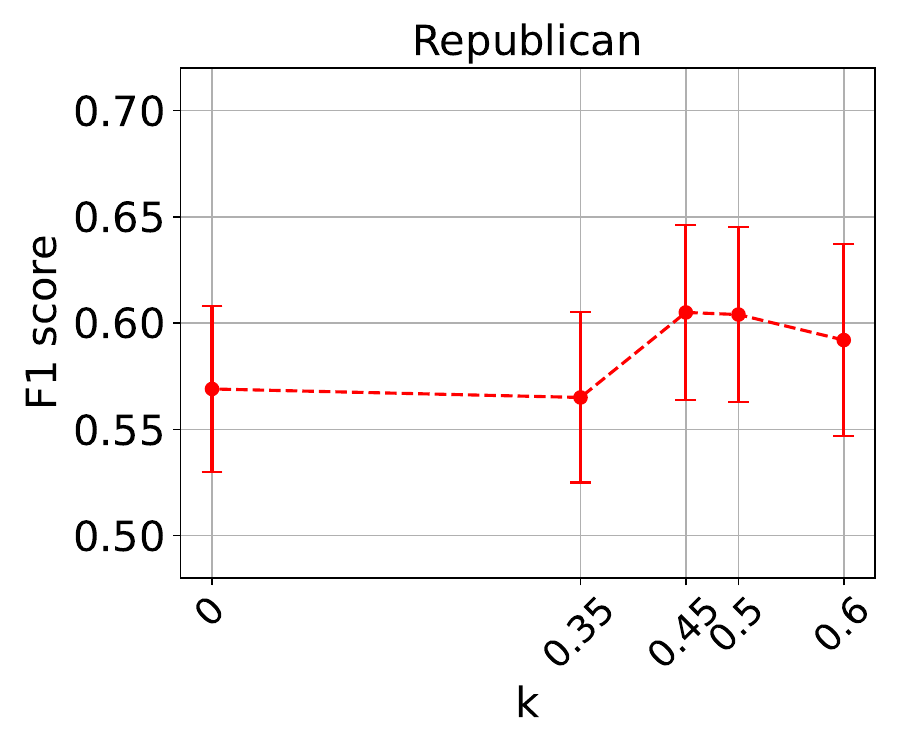}
    \caption{Group-level model performance at different $k$ values in \Dv. The error bars indicate 95\% confidence interval.}
    \label{fig:f1_v}
\end{figure*}

\begin{figure}[htb]
    \centering
    \includegraphics[width=0.5\textwidth]{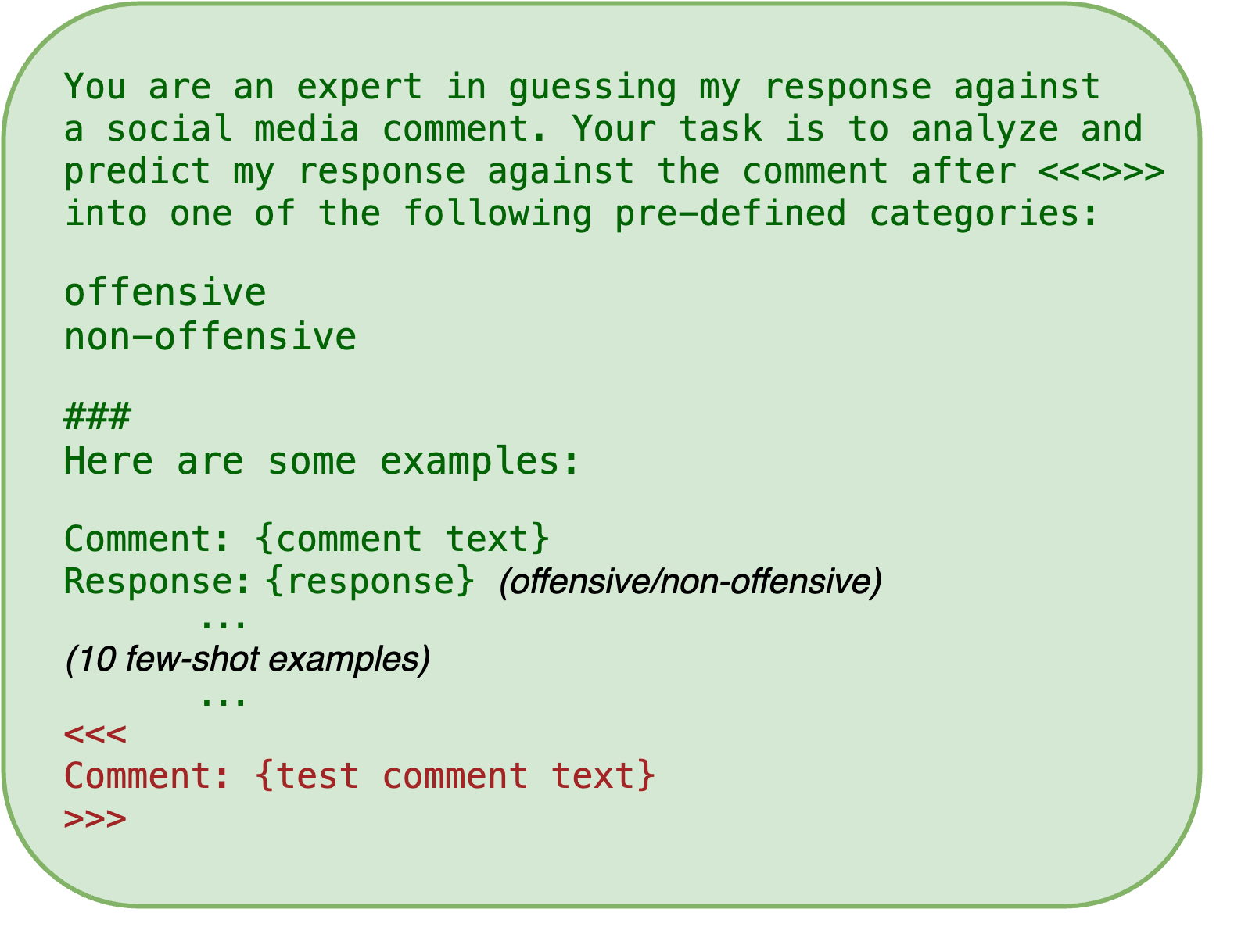}
    \caption{Prompt designed for ARTICLE.}
    \label{fig:prompt}
\end{figure}

We propose \texttt{\textbf{ARTICLE}} (\textbf{A}nnotator \textbf{R}eliability \textbf{T}hrough \textbf{I}n-\textbf{C}ontext \textbf{Le}arning) -- a two-step framework (Figure \ref{fig:article}) to identify reliable annotators and model the perception of offense for different political groups. In the fist step, we identify the annotators who exhibit inconsistency in labeling and remove them from the dataset. In the second step, based on the aggregated responses of the consistent annotators, we model the group-level perception of offense.

\subsection{Step 1: Identifying Inconsistent Annotators}
We hypothesize that annotators who show inconsistent annotation patterns are difficult to model. We individually model each annotator using a state-of-the-art LLM, \texttt{Mistral-7B-instruct} \cite{jiang2023mistral}, and utilize the model's performance (ease of modeling) as a proxy for the annotator's consistency. For each annotator, we randomly split their annotations into two sets -- the first set (training set) contains 10 data points, and the second (test set) contains the rest. Using the training set as in-context learning (ICL) \cite{dong2022survey, min2022rethinking} examples, we prompt \texttt{Mistral-7B-instruct} to predict the labels for the test set. The detailed prompt can be found in Figure \ref{fig:prompt}. Then, we compute the F1-score to evaluate the model's performance. A high F1 score indicates the annotator is easy to model and, hence, consistent, and a low score indicates the opposite. We define a hyperparameter ($k$) that acts as a threshold. If, for a given annotator, the F1-score is less than $k$, we mark them as inconsistent and remove them from the dataset.

\subsection{Step 2: Modeling Group-level Perception}
After removing the inconsistent annotators from all political groups, we recompute the aggregate labels for each group. We again use ICL to model the group-level perception of offense. For each group, we construct a training set using 70\% of the data. The rest is used for testing. For each test instance, we randomly sample 15 examples from the training set and use them as in-context examples. The same \texttt{Mistral-7B-instruct} model is used in this step.

\section{Experimental Setup}

\subsection{Datasets}
\begin{table}[htb]

  \centering
  \begin{tabular}{l c c}
    \hline
    Political Leaning & \Dtr & \Dv\\
    \hline
    Democrat & 43\% & 34\%\\
    Republican & 28\% & 36\%\\
    Independent & 29\% & 30\%\\
    \hline
  \end{tabular}
  \vspace{.2cm}
\caption{Distribution of political leanings of the annotators in \Dtr~ and \Dv.}  

\label{tab:dataset_stat}
\end{table} 

We consider two datasets on web toxicity: \Dtr~and \Dv. \Dtr~contains 107,620 comments from multiple social web platforms (Twitter, Reddit, and 4chan) collectively annotated by 17,280 annotators. We sample 20,000 comments from \Dtr~for our experiments ensuring that each set of 20 comments is annotated by the same five annotators, thereby retaining the structure of the original dataset. \Dv~includes 2,338 YouTube comments annotated by 726 annotators. Both datasets include annotators from diverse political backgrounds with at least 28\% (Table \ref{tab:dataset_stat}) representation from each major political affiliation -- Democrats, Republicans, and Independents.  In both datasets, comments are rated on a five-point scale of toxicity. To avoid rare classes, we convert these categories into binary labels. The lowest two toxic categories are mapped to \textit{non-offensive} class, and the rest are mapped to \textit{offensive} class.

\subsection{Models}
We primarily use \texttt{Mistra-7B-instruct} for the proposed framework; however, we have conducted experiments with \texttt{Llama3-8B-instruct} \cite{touvron2023llama} and \texttt{GPT-3.5-turbo} \cite{openai} to ensure the stability of the method.

\subsection{Computing Environment}
We run all our experiments in a Google Colab (pro+) environment with a single A100 GPU (40 GB) and 52 GB RAM.

\section{Inconsistent Annotation Examples}

Before we discuss our results and findings, we make a short detour to highlight why consistency matters through a few illustrative examples. We present annotations from a few inconsistent annotators as identified by \texttt{ARTICLE}. For instance, in the first example, both the comments express that the government should not control abortion; however, the annotator labeled them differently. In the second example, the annotator labeled a seemingly harmless comment offensive; however, assigned a non-offensive label to a comment containing a racial slur. 

\noindent\rule{\textwidth}{1pt}

\textbf{EXAMPLE ANNOTATOR \#1}

\raggedright{\noindent
\textbf{Comment:} \texttt{Abortion should be between the woman and God. I do agree the government should not control it. I'm not saying I agree with Abortion what so ever but the rioting and stuff is just ridiculous.}
}\\
\textbf{Label:} \textit{offensive}

\textbf{Comment:} \texttt{The `state' has no business deciding what women can or cannot do to their bodies. The fact that so many in the USA feel that abortion should be banned speaks more to a moronic pandering to the right wing religious fringe crazies who want to impose THEIR beliefs on everyone.}
\\
\textbf{Label:} \textit{non-offensive}
\noindent\rule{\textwidth}{1pt}

\textbf{EXAMPLE ANNOTATOR \#2}

\textbf{Comment:} \texttt{Amazing to see some of our ex students still continuing on their passion for the arts}
\\
\textbf{Label:} \textit{offensive}

\textbf{Comment:} \texttt{yea lil ass went Mia that boy a hog. n\censor{i}gg\censor{a}r finna average 3 assists a game a piece}
\\
\textbf{Label:} \textit{non-offensive}
\noindent\rule{\textwidth}{1pt}

\textbf{EXAMPLE ANNOTATOR \#3}

\textbf{Comment:} \texttt{If I Was A Cruddy n\censor{i}gg\censor{a}r Like I USED To Be And I Knew A Girl With An OnlyFans Account. I’d Cyber Pimp That Hoe}
\\
\textbf{Label:} \textit{non-offensive}

\textbf{Comment:} \texttt{Omfg is it like engagement season or something??? Like everyone is getting engaged right now. Wtf}
\\
\textbf{Label:} \textit{offensive}
\noindent\rule{\textwidth}{1pt}

\textbf{EXAMPLE ANNOTATOR \#4}

\textbf{Comment:} \texttt{Oh you wanna be part of my business venture? You can help fill the twinkies with c\censor{u}m}
\\
\textbf{Label:} \textit{non-offensive}

\textbf{Comment:} \texttt{Can’t wait to see you guys}
\\
\textbf{Label:} \textit{offensive}\\
\noindent\rule{\textwidth}{1pt}
\justifying

\section{Evaluation}

\subsection{Modeling Performance}

We evaluate the proposed framework on \Dtr~ and \Dv. In each dataset, we model the perception of offense for each political group: Democrat, Republican, and Independent. As mentioned earlier, our framework requires setting a value for the hyperparameter $k$. To study the impact of $k$, we run experiments for the following values of $k: \{0, 0.35, 0.45, 0.5, 0.6\}$. The case $k=0$ serves as the baseline where we do not remove any annotators from the dataset. Figures \ref{fig:f1_tr} and \ref{fig:f1_v} illustrate the performance (F1-score on the test set) at various values of $k$ for \Dtr and \Dv, respectively. In general, in both the datasets, across all political groups, we observe an upward trend in the F1-score as the value of $k$ increases with noticeable fluctuations for Independents. In almost all instances, the F1-score achieved with $k=0.45$ surpassed the baseline performance, suggesting the effectiveness of the proposed method. We also note
for most cases with $k>0.5$, the performance either plateaus or declines slightly. It suggests that while increasing $k$ generally improves model performance up to a point, there may be a threshold beyond which further increase in $k$ does not yield additional benefits and might even be detrimental.

\subsection{Data Loss}

\begin{figure*}[htb]
\centering
\begin{subfigure}{0.45\textwidth}
  \centering
  \includegraphics[width=0.9\linewidth]{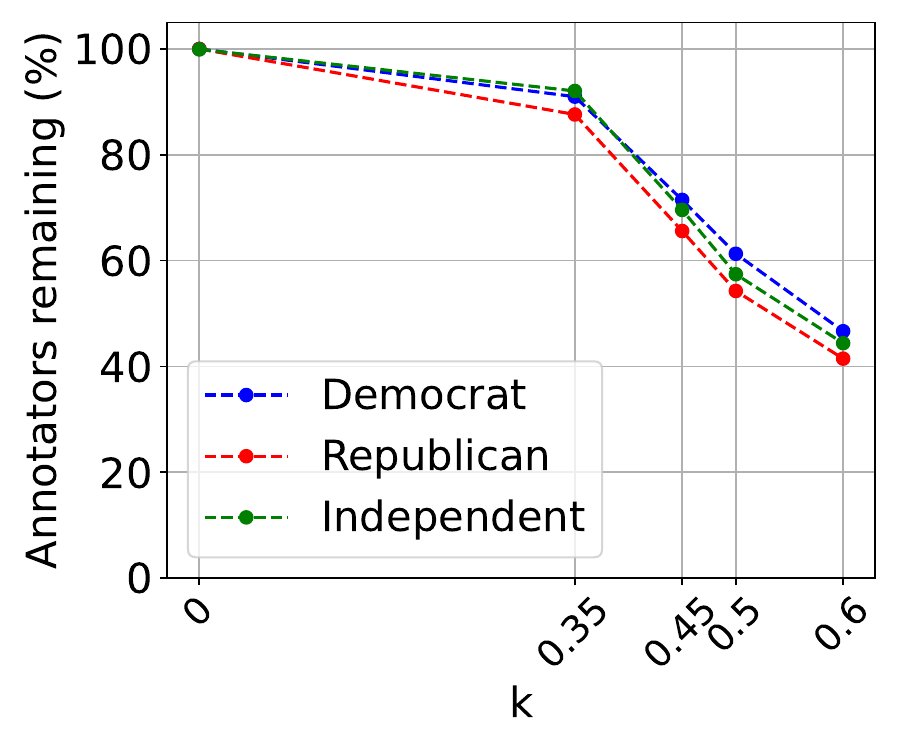}
  \caption{\Dtr}
  \label{fig:ann_tr}
\end{subfigure}%
\begin{subfigure}{0.45\textwidth}
  \centering
  \includegraphics[width=0.9\linewidth]{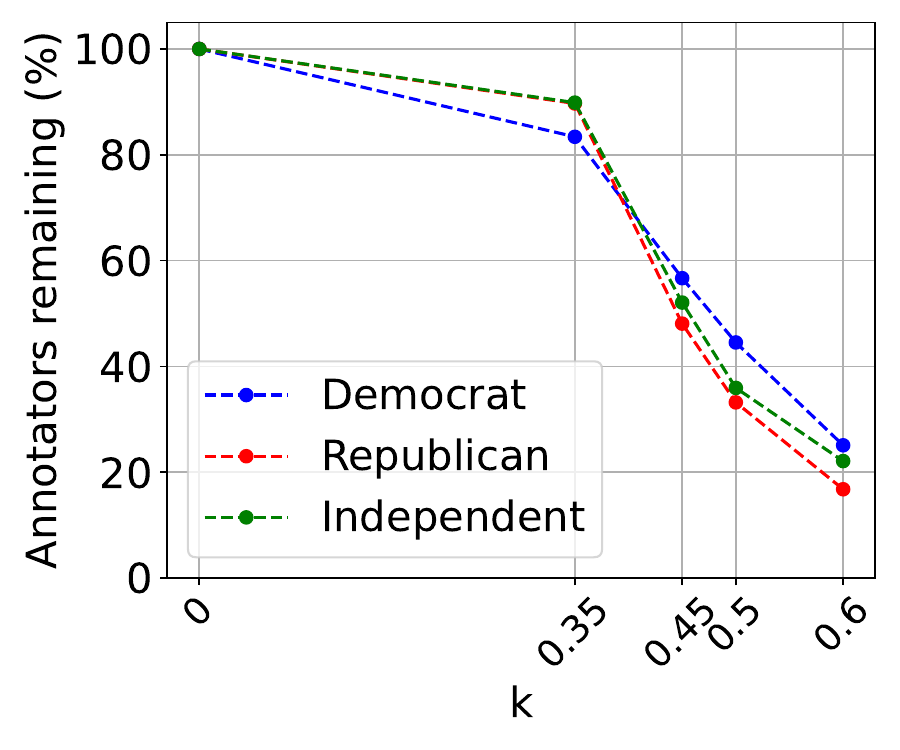}
  \caption{\Dv}
  \label{fig:ann_v}
\end{subfigure}

\begin{subfigure}{0.45\textwidth}
  \centering
  \includegraphics[width=0.9\linewidth]{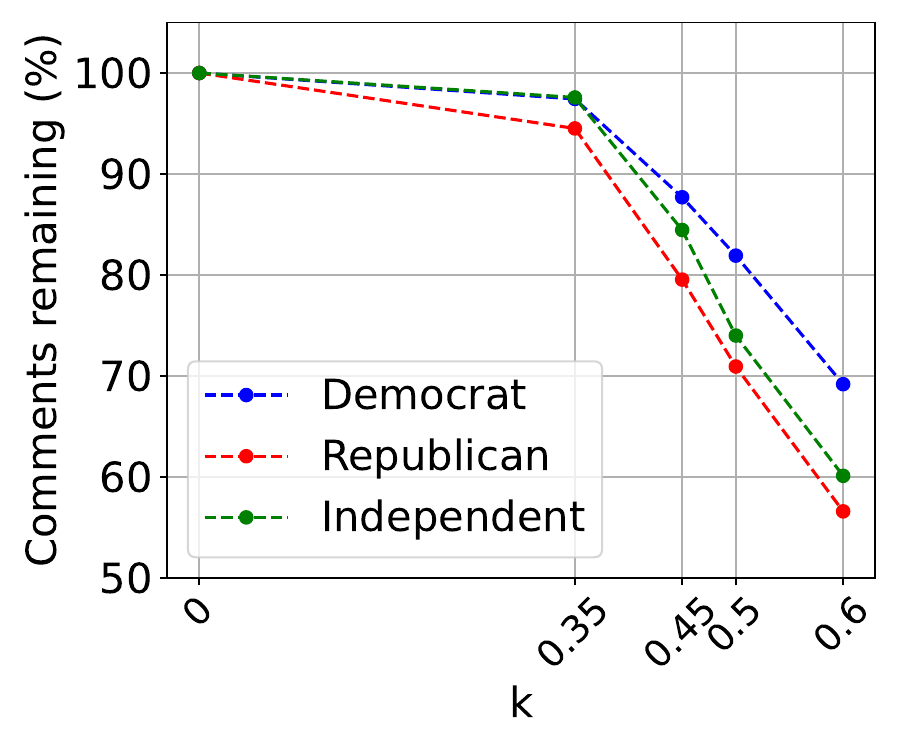}
  \caption{\Dtr}
  \label{fig:comm_tr}
\end{subfigure}%
\begin{subfigure}{0.45\textwidth}
  \centering
  \includegraphics[width=0.9\linewidth]{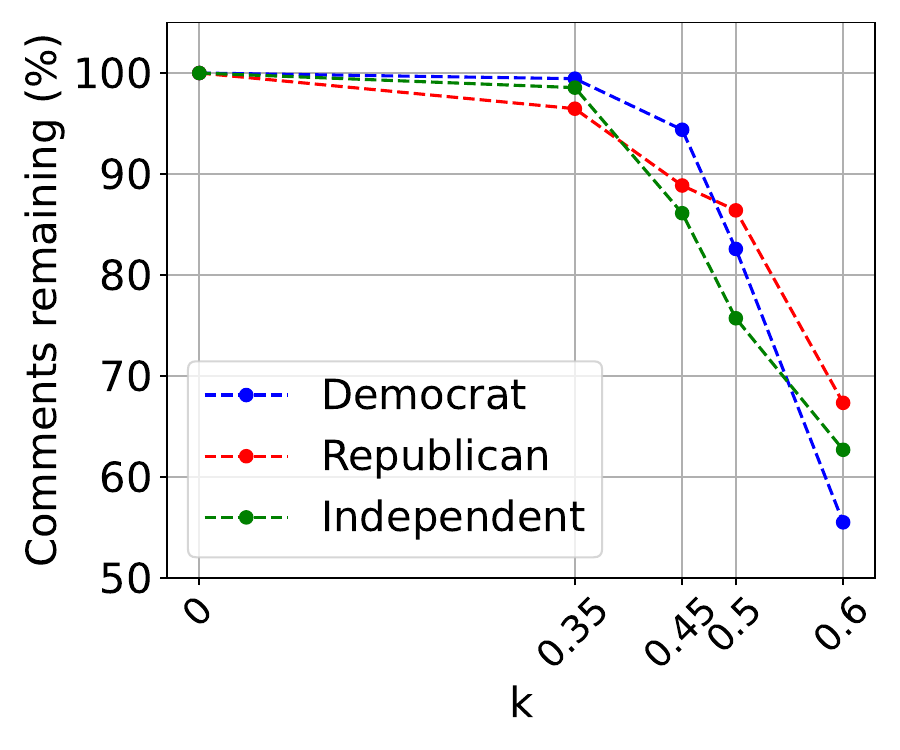}
  \caption{\Dv}
  \label{fig:comm_v}
\end{subfigure}
\caption{Percentage of annotators and comments remaining at various value of $k$ in \Dtr~ and \Dv.}
\label{fig:test}
\end{figure*}

While increasing $k$ improves modeling performance, the annotations lost in this process merit investigation. We first compute the percentage of the annotators remaining at various values of $k$. From Figures \ref{fig:ann_tr} and \ref{fig:ann_v}, we note that \Dv~undergoes a sharper decline in annotators compared to \Dtr. However, at $k=0.45$, we still retain the majority ($\sim$ 70\% in \Dtr~ and $\sim$ 55\% in \Dv) of the annotators in both datasets, with Democrats generally showing the highest retention rates.

Next, we focus on the number of comments remaining as we increase $k$. We again compute this at group level for \Dtr~(Figure \ref{fig:comm_tr}) and \Dv ~(Figure \ref{fig:comm_v}). \texttt{ARTICLE} at $k=0.45$, retains more than 80\% of the comments in both datasets.

\subsection{Comparison with CT}

\begin{table}[htb]
    \centering
    \begin{tabular}{l c c}
    \hline
    Political Leaning & CT (\textit{WQS} $\ge 0.6$) & \texttt{ARTICLE} $(k \ge 0.45)$ \\
    \hline
    Democrat & $0.669 \pm 0.016$ & $0.696 \pm 0.015$ \\
    Republican & $0.642 \pm 0.018$ & $0.671 \pm 0.017$ \\
    Independent & $0.665 \pm 0.018$ & $0.696 \pm 0.017$ \\
    \hline
    \end{tabular}
    \vspace{.2cm}
        \caption{Group-level modeling performance (F1-score on test set) comparison between \texttt{ARTICLE} and CT in \Dtr. The results are computed over five runs with different random seeds.}
        \label{ct_vs_article}
\end{table}

\begin{table}[htb]

    \centering
    \begin{tabular}{l c c}
    \hline
    Political Leaning & CT (\textit{WQS} $\ge 0.7$) & \texttt{ARTICLE} $(k \ge 0.45)$ \\
    \hline
    Democrat & $0.449 \pm 0.036$ & $0.616 \pm 0.041$ \\
    Republican & $0.435 \pm 0.032$ & $0.605 \pm 0.042$ \\
    Independent & $0.453 \pm 0.036$ & $0.557 \pm 0.042$ \\
    \hline
    \end{tabular}
    \vspace{.2cm}
        \caption{Group-level modeling performance (F1-score on test set) comparison between \texttt{ARTICLE} and CT in \Dv. The results are computed over five runs with different random seeds.}
        \label{ct_vs_article_voiced}
\end{table}

We compare our framework with CT, a well-known method of estimating the quality of annotations \cite{dumitrache2018crowdtruth}. CT computes multiple metrics on the annotated dataset, among which \textit{WQS} measures the quality of the annotators. The value of \textit{WQS} ranges between $[0,1]$. We consider annotators who score more than (or equal to) a specific \textit{WQS} value and model their aggregated annotations following the second step of \texttt{ARTICLE}. Using \Dtr, we choose $WQS = 0.6$, as in this setting, CT retains a similar percentage ($\sim 70\%$) of annotators to \texttt{ARTICLE} ($k=0.45$). Table \ref{ct_vs_article} shows that \texttt{ARTICLE} outperforms CT across all groups. The results for \Dv~ are presented in Table \ref{ct_vs_article_voiced}. Here, too, we notice a significant performance improvement with \texttt{ARTICLE} over CT.

\begin{figure*}[htb]
    \centering
    \includegraphics[width=0.24\textwidth]{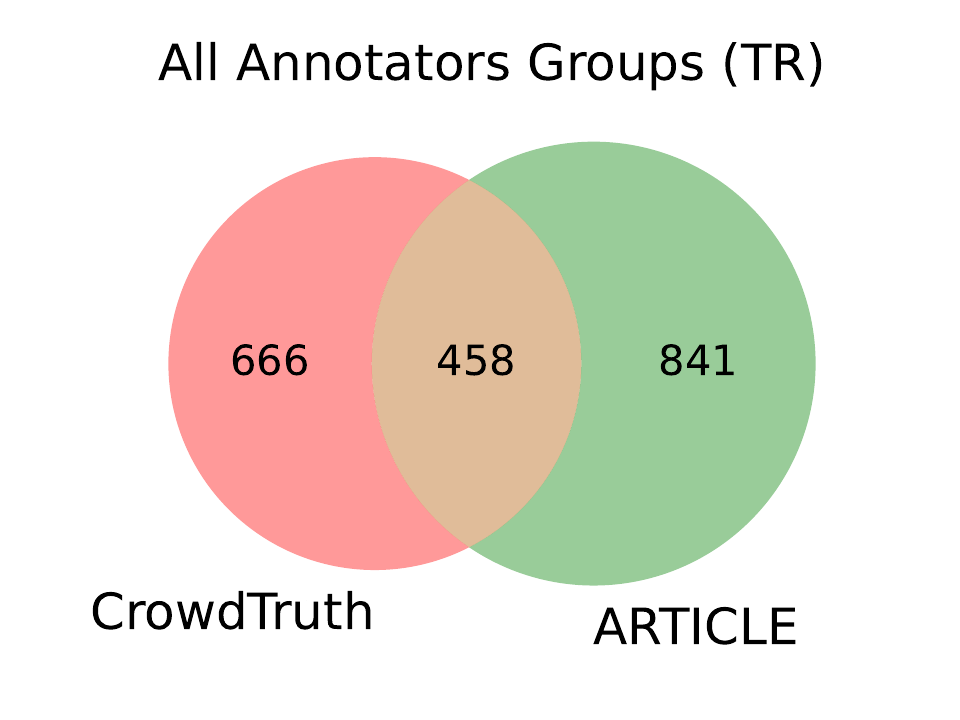}
    \includegraphics[width=0.24\textwidth]{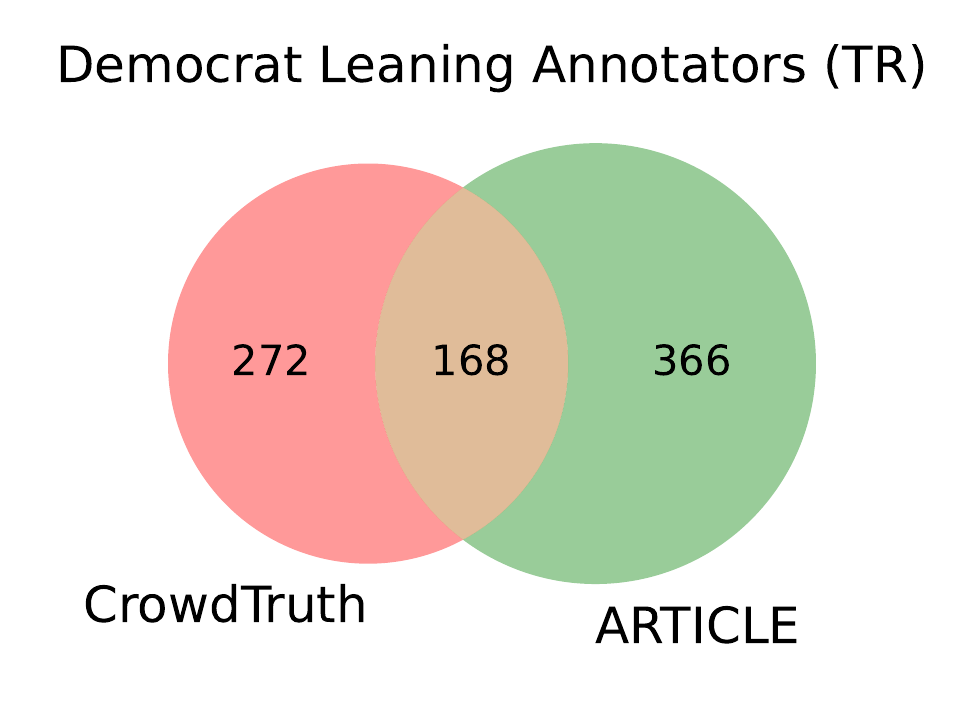} 
    \includegraphics[width=0.24\textwidth]{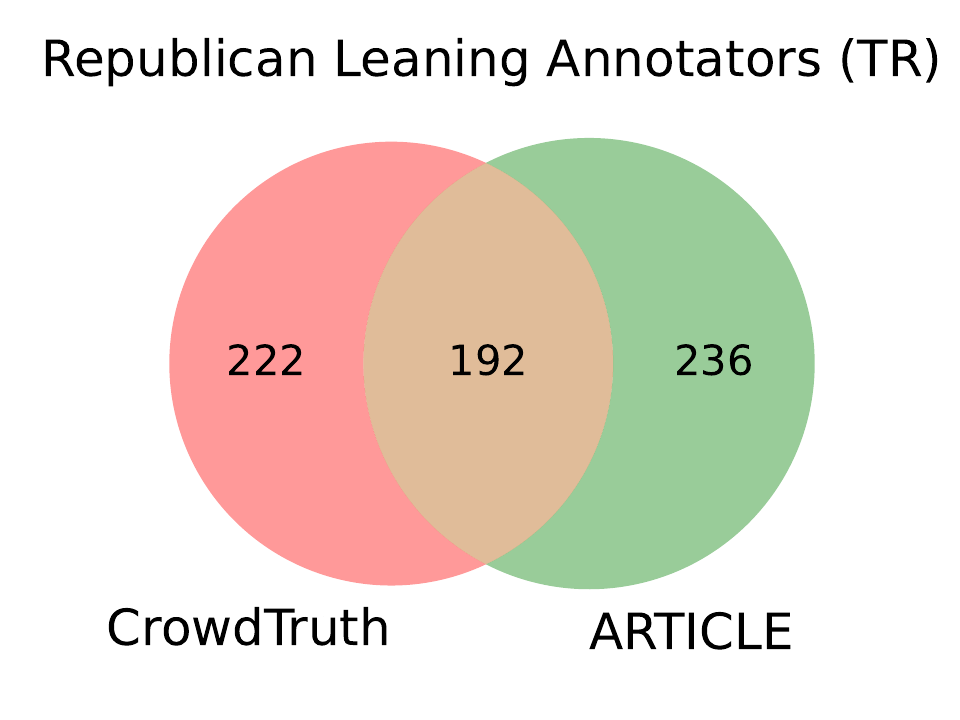}
    \includegraphics[width=0.24\textwidth]{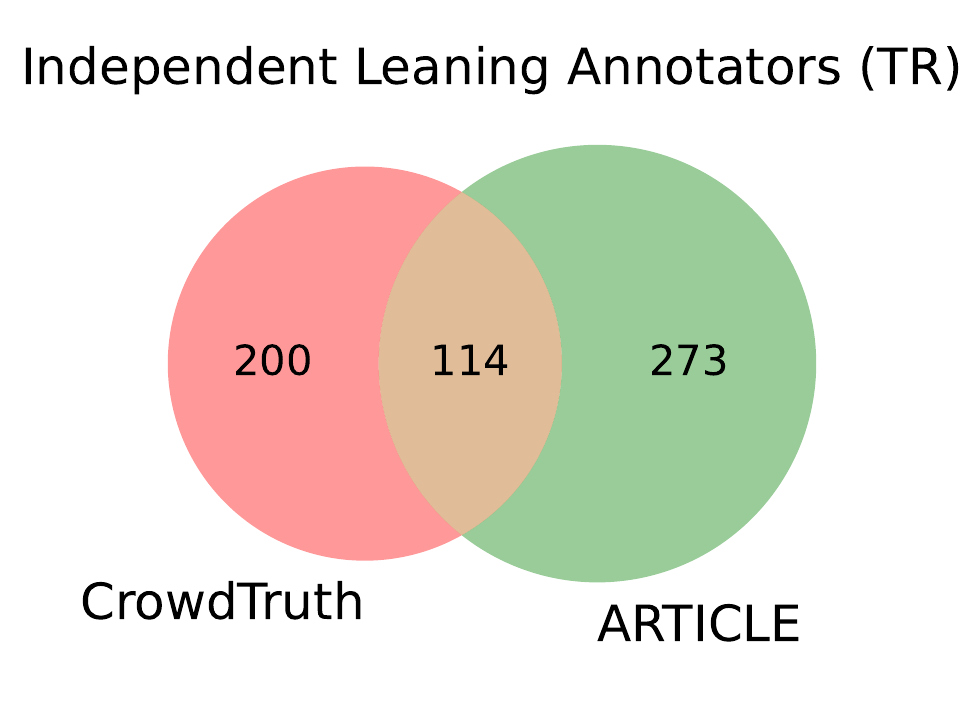}
    \caption{Annotators that are identified as unreliable based on CT and \texttt{ARTICLE} scores for \Dtr. The last three Figures show the same inconsistent annotators broken down by their political leaning. For \Dtr, CT (\textit{WQS} $\ge 0.6$) and \texttt{ARTICLE} $(k \ge 0.45)$. }
    \label{fig:ap-tr_venn_all}
\end{figure*}

\begin{figure*}[htb]
    \centering
    \includegraphics[width=0.24\textwidth]{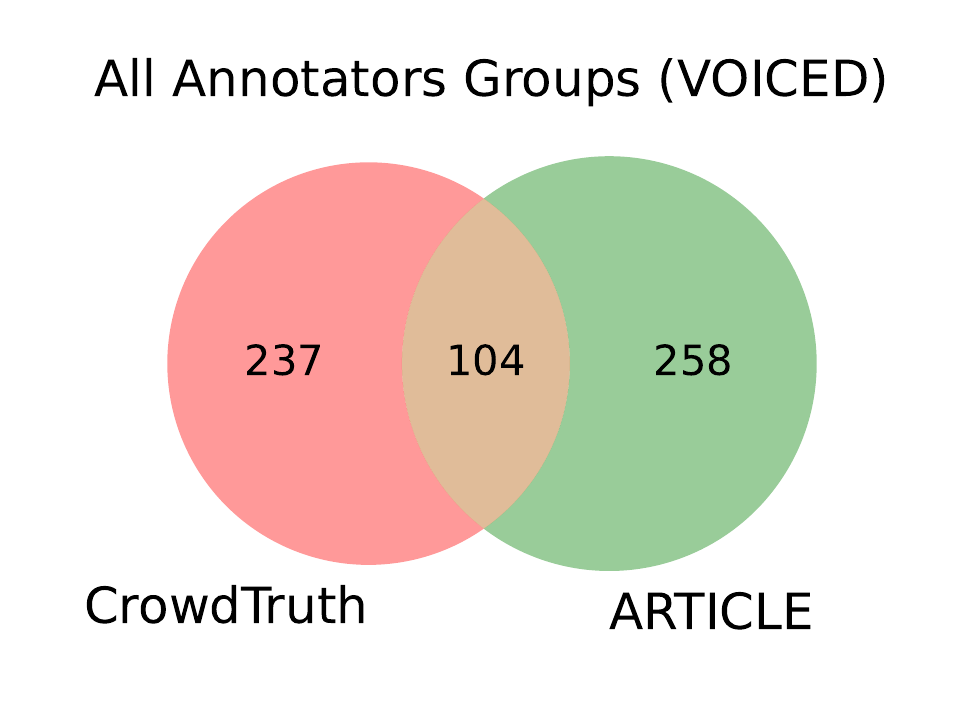}
    \includegraphics[width=0.24\textwidth]{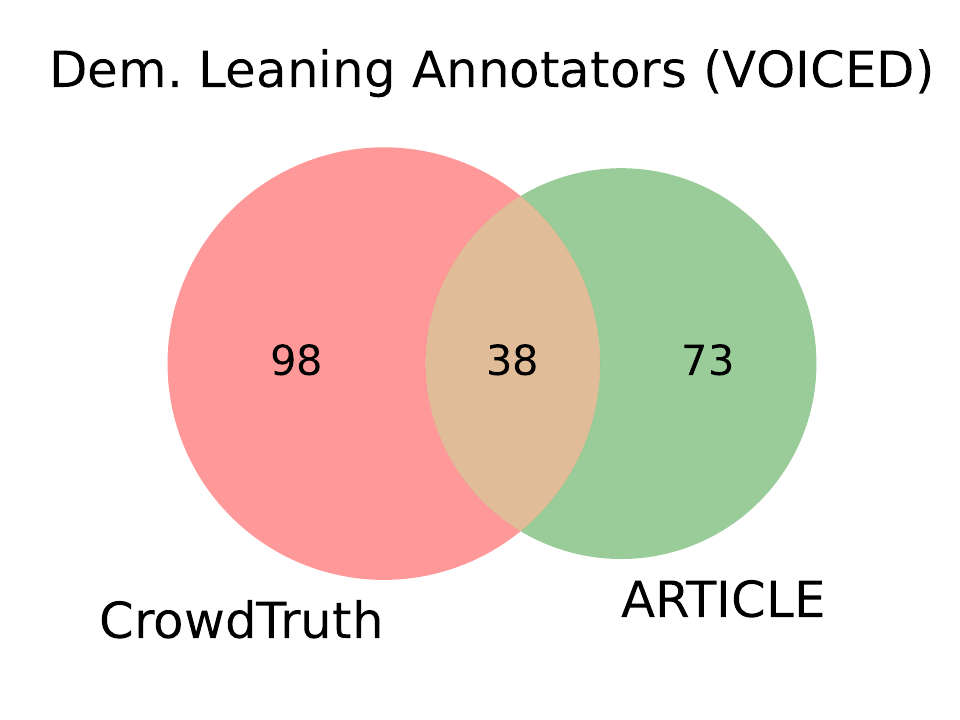}   
    \includegraphics[width=0.24\textwidth]{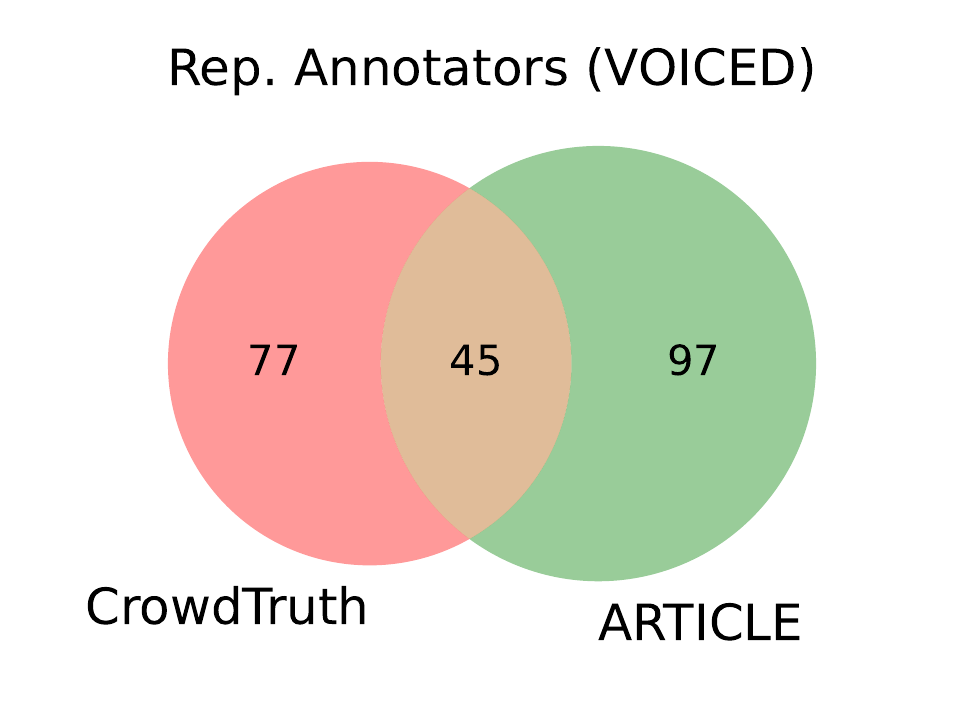}
    \includegraphics[width=0.24\textwidth]{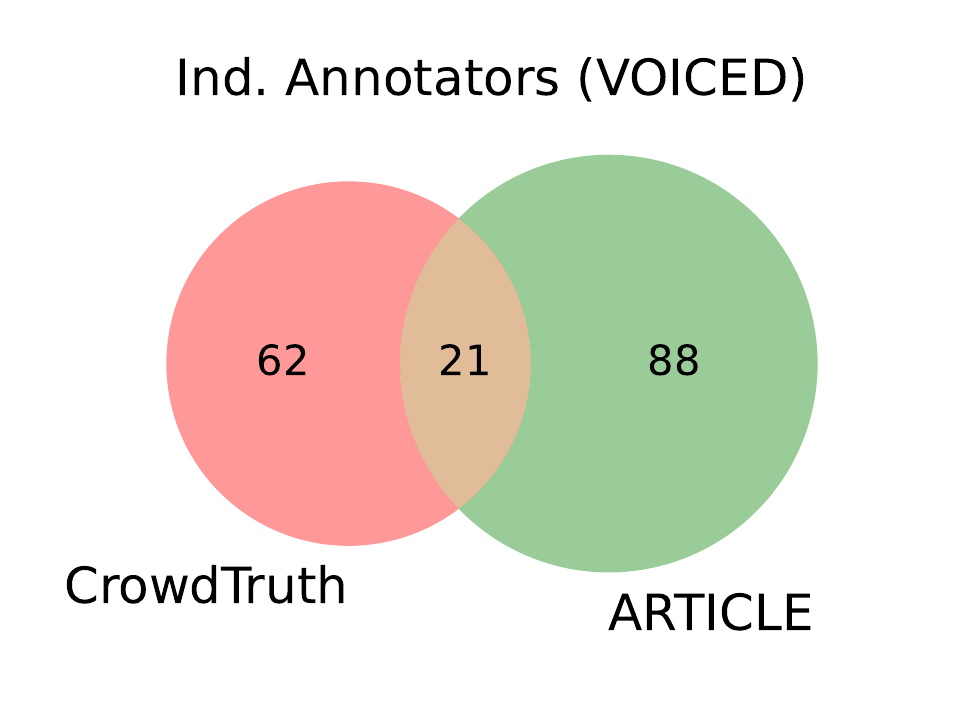}
    \caption{Annotators that are identified as unreliable based on CT and \texttt{ARTICLE} scores for \Dv. The last three Figures show the same inconsistent annotators broken down by their political leaning. For \Dv, CT (\textit{WQS} $\ge 0.86$) and \texttt{ARTICLE} $(k \ge 0.46)$.}
    \label{fig:ap-voiced_venn_all}
\end{figure*}

We further investigate the overlap between \texttt{ARTICLE} and CT. Figures \ref{fig:ap-tr_venn_all} and \ref{fig:ap-voiced_venn_all} show the venn diagram between the low-quality annotators identified by the two methods in \Dtr~ and \Dv. We observe that while there is a substantial overlap between the two methods, there are annotators who are flagged as low-quality by one but not by the other. This suggests that these methods measure slightly different aspects of the annotation quality, and future work should explore ways to combine them in a single pipeline.

\subsection{Stability across LLMs}
\begin{table}[htb]
\centering
\begin{tabular}{|l|c|c|c|}
\hline
 & \texttt{Mistral} & \texttt{Llama3} & CT \\ \hline
\texttt{Mistral} & - & 0.60 & 0.35 \\ \hline
\texttt{Llama3} & 0.60 & - & 0.40 \\ \hline
CT & 0.35 & 0.40 & - \\ \hline
\end{tabular}
\vspace{.2cm}
\caption{Jaccard similarities between inconsistent annotators identified by \texttt{ARTICLE} using different LLMs in \Dtr. It also includes similarities between each LLM and CT. Due to resource limitations, GPT was not used for this dataset.}
\label{tab:jacc_tr}
\end{table}

\begin{table}[ht]
\centering
\begin{tabular}{|l|c|c|c|c|}
\hline
 & \texttt{Mistral} & \texttt{Llama3} & \texttt{GPT} & CT \\ \hline
\texttt{Mistral} & - & 0.68 & 0.65 & 0.18 \\ \hline
\texttt{Llama3} & 0.68 & - & 0.65 & 0.16 \\ \hline
\texttt{GPT} & 0.65 & 0.65 & - & 0.20 \\ \hline
CT & 0.18 & 0.16 & 0.20 & - \\ \hline
\end{tabular}
\vspace{.2cm}
\caption{Jaccard similarities between inconsistent annotators identified by \texttt{ARTICLE} using different LLMs in \Dv. It also includes similarities between each LLM and CT.}
\label{tab:jacc_v}
\end{table}

Beyond \texttt{Mistral-7B-instruct}, we study the robustness of the \texttt{ARTICLE} framework across multiple LLMs. We consider two additional models: \texttt{Llama3-8B-instruct} \cite{touvron2023llama} (open-sourced) and \texttt{GPT-3.5-turbo} \cite{openai} (v. 0125, proprietary). To study the stability of our framework, we look at the overlap between the inconsistent annotators found by different LLMs. More precisely, we compute the Jaccard similarity between the sets of inconsistent annotators identified using a pair of LLMs. To ensure a fair evaluation, for each LLM, we consider the annotators who score less than the median as the inconsistent annotators. Table \ref{tab:jacc_tr} and \ref{tab:jacc_v} present the Jaccard similarities among the LLMs pairs in \Dtr and \Dv, respectively. We find a substantial ($\ge0.60$) similarity between every pair of LLM in both datasets, suggesting the stability of the framework. We also report the similarity between the annotators found by different LLMs with CT. The similarities between each LLM and CT are much lower ($\le0.40$) than between any two LLMs. This result indicates that CT does not identify many inconsistent annotators as poor-quality annotators. On the other hand, \texttt{ARTICLE} does not remove many of the annotators deemed unreliable by CT.

\section{Conclusion}
We introduce \texttt{ARTICLE}, a novel framework for estimating annotator quality through self-consistency. Our approach marks a significant shift from traditional outlier-based methods. Evaluations across two offensive speech datasets demonstrate that \texttt{ARTICLE} effectively identifies reliable annotators while preserving unique, self-consistent viewpoints that might be overlooked. Furthermore, the consistent performance of \texttt{ARTICLE} across multiple language models highlights its robustness. Focusing on self-consistency reduces the dependence on larger annotator pools, potentially lowering costs and increasing the feasibility of deploying quality control mechanisms in annotation tasks. The ongoing development of \texttt{ARTICLE} aims to enhance our understanding and management of the subjective nature of annotation, paving the way for more reliable and inclusive data collection methods.

\section*{Limitations}

While \texttt{ARTICLE} introduces a promising approach to annotator quality assessment, several limitations warrant further investigation:

\subsection{Model Bias} Reliance on LLMs for evaluating self-consistency could introduce biases inherent to these models~\cite{bommasani2022picking, IJCAI2024RabbitHole}. These biases may affect the framework's ability to accurately estimate the quality of annotations, especially in contexts involving linguistic or cultural nuances that LLMs might not fully capture.

\subsection{Handling Justified Disagreement:} \texttt{ARTICLE} currently lacks a robust mechanism to distinguish between justified disagreements and genuine inconsistencies in annotations which merits deeper exploration.

\subsection{Generalizability Across Domains:} While tested on datasets involving offensive speech, the generalizability of the framework to other types of annotation tasks, such as medical image annotation or legal document analysis, remains unverified. Different domains may present unique challenges that require adaptations of the framework.

\subsection{Dependency on Annotation Volume:} The effectiveness of \texttt{ARTICLE} is constrained by the volume of data available for each annotator. In scenarios where annotators contribute a low number of annotations, the assessment of self-consistency could be less reliable.

\section*{Ethics Statement}

\texttt{ARTICLE}'s approach to annotation quality assessment through self-consistent intends to help mitigate potential biases towards minor perspectives in NLP systems. In this work, we used two publicly available datasets referenced in the paper. No new data collection has been carried out as part of this work. The datasets used do not reveal any identifiable information about the annotators.





\end{document}